# On the Combinability of Evidence in the Dempster-Shafer Theory

*Lotfi A. Zadeh†*

and

*Anca Ralescu‡*

## Abstract

In the current versions of the Dempster-Shafer theory, the only essential restriction on the validity of the rule of combination is that the sources of evidence must be statistically independent. Under this assumption, it is permissible to apply the Dempster-Shafer rule to two or more distinct probability distributions.

An essential step in the Dempster-Shafer rule of combination of evidence is that of normalization. The validity of normalization is open to question, particularly in application to probability distributions (Zadeh, 1976). At this juncture, the validity of normalization is a controversial issue.

In this paper, we construct a relational model for the Dempster-Shafer theory which greatly simplifies the derivation of its main results and cast considerable light on the validity of the rule of combination. The relational model is augmented with what is called the ball-box analogy, yielding an intuitively simple way of visualizing the concepts of belief and plausibility.

---

† Computer Science Division, University of California, Berkeley, CA 94720.

‡ Department of Computer Science, University of Cincinnati, Cincinnati, OH 45221.

Research supported in part by NSF Grant DCR-8513139, NASA Grant NCC-2-275 and NESC Contract N00039-84-C-0243.

In the relational model, a multivalued mapping is represented as a second-order relation in which the attributes are *granular*, that is, set-valued. In this model, a Dempster-Shafer distribution is a granular distribution which may be interpreted as a summary of the *parent* relation, that is, the relation which represents the multivalued mapping from individuals to their attributes. Given a set-valued query, Q, the number of individuals whose attribute is certain to satisfy Q is the *necessity* of Q, while the number of individuals whose attribute may possibly satisfy Q is the *possibility* of Q. Necessity and possibility are the counterparts of belief and plausibility in the Dempster-Shafer theory (Zadeh, 1979a, 1986).

In the relational model, two distinct sources of evidence are represented as two distinct columns for a single attribute in the parent relation. A parent relation is said to be *conflict-free* if the intersection of set-valued entries in the two columns is non-empty for each individual. From this model, it follows that, in order to be combinable, the sources of evidence must have a common parent relation which is conflict-free.

More generally, let $G = \{(A_1, p_1), ..., (A_m, p_m)\}$ and $H = \{(B_1, q_1), ..., (B_n, q_n)\}$ be two granular distributions in which $p_i$, $i = 1, ..., m$ is the relative count of individuals in the parent relation whose set-valued attribute is $A_i$ in one column, and $q_j$, $j = 1, ..., n$, is the relative count of individuals whose set-valued attribute is $B_j$ in another column, with the understanding that the columns in question represent two distinct sources of evidence. Then, a sufficient condition for noncombinality is that there is a granule $A_i$ in G which is disjoint from all granules in H, or vice-versa. This condition, however, is not necessary, as is demonstrated by the following example: $G = \{(A_1, 2/3), (A_2, 1/3)\}$. $H = \{(A_1, 1/3), (A_2, 2/3)\}$, in which $A_1$ and $A_2$ are disjoint. In this case, it is evident that there does not exist a parent relation which is conflict-free.

A necessary and sufficient condition for noncombinability which is computationally much more efficient than a direct test based on the definition of noncombinability, is formulated for the case where the granules in each distribution are disjoint. In addition, the ball-box analogy is employed to derive some of the basic results of the Dempster-Shafer theory, and to extend it to cases in which the sources have unequal credibilities.

## References and Related Publications